\documentclass[10pt,conference,a4paper]{IEEEtran}

\ifCLASSINFOpdf
\usepackage[pdftex]{graphicx}
\else
\fi

\usepackage{ifpdf}
\usepackage{cite}
\usepackage{amsmath}
\usepackage{latexsym}  
\usepackage{booktabs}
\usepackage{algorithmic}
\usepackage{array}
\usepackage{fixltx2e}
\usepackage{stfloats}
\usepackage{url}
\usepackage{epsfig}
\usepackage{graphicx}
\usepackage{amssymb}
\usepackage{multirow}
\usepackage{xspace}

\usepackage{adjustbox}
\usepackage{multirow}
\usepackage{blindtext}
\usepackage{graphicx}
\usepackage{caption}

\begin{document}
%
\title{Exploiting Style and Attention in Real-World Super-Resolution}

\author{\IEEEauthorblockN{Xin Ma$^{1,2}$, Yi Li$^{1,2}$, Huaibo Huang$^{1,2}$, Mandi Luo$^{1,2}$, Ran He$^{1,2}$}
\IEEEauthorblockA{{$^1$}School of Artificial Intelligence, University of Chinese Academy of Sciences
$^{2}$NLPR$\;\&\;$CEBSIT, CASIA \quad\\
{\tt\small \{xin.ma, yi.li, huaibo.huang\}@cripac.ia.ac.cn, luomandi2019@ia.ac.cn, rhe@nlpr.ia.ac.cn}}
}

\maketitle

\begin{abstract}
Real-world image super-resolution (SR) is a challenging image translation problem. Low-resolution (LR) images are often generated
 by various unknown transformations rather than by applying simple bilinear down-sampling on high-resolution (HR) images. 
 To address this issue, this paper proposes a novel pipeline which exploits style and attention mechanism in real-world SR. Our pipeline consists of a style Variational Autoencoder (styleVAE) and a SR network incorporated with attention mechanism. To get real-world-like low-quality images paired with the HR images, we design the styleVAE to transfer the complex nuisance factors in real-world LR images to the generated LR images. We also use mutual information estimation (MI) to get better style information. For our SR network, we firstly propose a global attention residual block to learn long-range dependencies in images. Then another local attention residual block is proposed to enforce the attention of SR network moving to local areas of images in which texture detail will be filled. It is worth noticing that styleVAE can be presented in a plug-and-play manner and thus can help to improve the generalization 
 and robustness of our SR method as well as other SR methods. Extensive experiments demonstrate that our method surpasses the state-of-the-art work, both quantitatively and qualitatively.
\end{abstract}

\section{Introduction}
Single image super-resolution (SISR) aims to infer a natural high-resolution (HR) image from the degraded low-resolution (LR) input.
SISR technology is widely used in a variety of computer vision tasks, e.g., security surveillance, medical image processing, and face recognition.

Recently, many deep learning based super-resolution (SR) methods have been greatly developed and achieved promising results. 
These methods are mostly trained on paired LR and HR images, while the LR images are usually obtained by performing a predefined degradation mode on the HR images, e.g., bicubic interpolation.
Different from them, we argue that it will hinder the generalization and robustness of an SR method if it is trained on man-made image pairs, especially when confronting real-world LR images.

\begin{figure}[ht]
    \centering
    \includegraphics[scale=0.35]{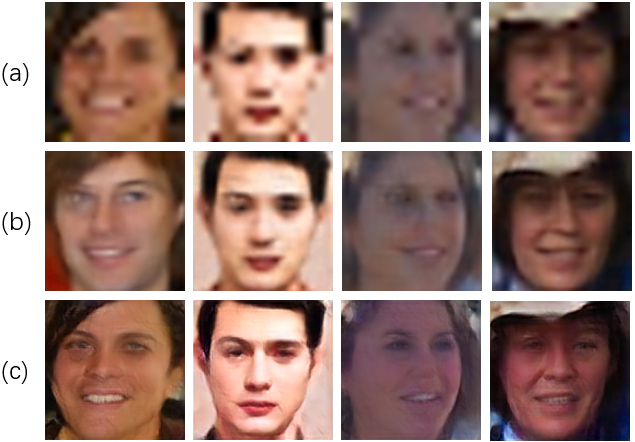}
    \caption{Comparison of our method with SRGAN \cite{ledig2017photo}. (a) Input real-world LR images. (b) Results of SRGAN ($8\times$). (c) Results of our method ($8\times$).}
    \label{Fig. 1}
\end{figure}

In fact, there is a huge difference between the LR images after bicubic interpolation and real-world LR images. 
There are various nuisance factors leading to image quality degeneration, e.g., motion blur, lens aberration and sensor noise.
Moreover, these nuisance factors are usually unknown and mixed up with each other, making the real-world SR task blind and thus stunning challenging.
The LR generated manually can only simulate limited patterns and methods trained on them inherently lack the ability of dealing with real-world SR issues. 
As presented in the second row of Fig. \ref{Fig. 1}. The performance of SRGAN drops dramatically when fed with real-world LR images and natural HR images can not be produced anymore.

Despite these, relatively less attention has been paid to the problem of how to super-resolve real-world LR images until now. 
Some generative adversarial network (GAN) based methods \cite{bulat2018learn, zhou2019kernel} have made some explorations, but the effects are primary and there is still much room for improvement.
A straightforward idea is to capture a large number of different resolution images simultaneously in the same scene. 
However, it is not a reasonable solution considering human and material constraints. 
In order to solve this problem in an efficient manner, we propose a generative network based on variational autoencoders (VAEs) to synthesize real-world-like LR images. In this paper, we conduct  experiments on face images (i.e., face hallucination), but the proposed method is also applicable to other image types.

The essential idea is derived from the separable property of image style and image content, which has been widely explored in image style transfer \cite{gatys2016image,huang2017arbitrary, kotovenko2019content}. 
It means that one can change the style of an image while preserving its content. 
Based on these previous researches, we propose to consider the fore-mentioned nuisance factors as a special case of image styles.
We then design styleVAE to transfer the complex nuisance factors in real-world LR images to generated LR images.
In this manner, real-world-like LR images as well as LR-HR pairs can be generated automatically. 
Furthermore, styleVAE can be presented as a plug-and-play component and can also be applied to existing SR methods to improve their generalization and robustness. 

In addition to styleVAE, we build an SR network for real-world super-resolution.
Most CNN-based SR methods impel element-wise features to be equal, which limits the ability of a deep neural network to represent different types of information, e.g. low-frequency and high-frequency information. The progress in biology inspires us with the observation that human visual perception systems follow the principle of global priority. Following this principle, our proposed SR network consists mainly of two modules. On the one hand, we develop a global attention residual block (GARB) to capture long-range dependency correlations, helping the SR network to focus on global topology information. On the other hand, we also introduce a local attention residual block (LARB) for better feature learning, which is essential to infer high-frequency information in images.

Our major contributions in this paper can be summarized as follows:
\begin{itemize}
\item We propose to generate real-world-like LR images by transfering various degradation modes in reality and obtain paired LR and HR images with a newly designed styleVAE. It is worth noticing that styleVAE can be presented in a plug-and-play manner, promoting the generalization and robustness of existing SR methods.
\item An SR network is developed for real-world super-resolution. There are two main components within the SR network. We propose GARB to adaptively rescale features via considering inter-dependencies among feature elements. At the same time, LARB is introduced to maximize the recovery of high-frequency information.
\item Extensive experiments on real-world LR images demonstrate that styleVAE effectively facilitates SR methods and the proposed SR network achieves state-of-the-art results in terms of visual and numerical quality.
\end{itemize}

\section{Related Work}
\textbf{Super-Resolution Neural Networks.} Last few years, machine learning has enjoyed another Renaissance based on deep learning. Some surprising results have been achieved by deep learning-based methods in the field of super-resolution. Dong \textit{et al.} \cite{dong2015image} firstly attempted to use deep convolutional neural networks (CNNs) called SRCNN. Kim \textit{et al.} \cite{kim2016accurate} designed a deeper network by using residual learning to get more network capacity. Johnson \textit{et al.} \cite{johnson2016perceptual} calculated mean square error (MSE) on feature maps to improve the perceptual quality of the output. Shi \textit{et al.} \cite{shi2016real} proposed an up-sample mode called sub-pixel convolution layer, which can super-resolve LR images in real-time. Ledig \textit{et al.} \cite{ledig2017photo} firstly adopted GAN-based networks \cite{goodfellow2014generative} to reconstruct photo-realistic images. To address the problem of lacking high-frequency textures,  Sajjadi \textit{et al.} \cite{sajjadi2017enhancenet} proposed a novel application for automating texture synthesis. Zhang \textit{et al.} \cite{zhang2018residual} proposed a novel residual dense network (RDN), which can make full use of the hierarchical features of LR images. Li \textit{et al.} \cite{li2018multi} proposed a novel multi-scale residual network to utilize image features fully. Li \textit{et al.} \cite{li2019feedback} exploited feedback mechanisms in SISR. 

\textbf{Face Super-Resolution.} Unlike general SISR, face super-resolution focuses more on face-specific information. The resolution of face image in training datasets is smaller, usually $16\times16$ or $32\times32$. Some previous methods make use of the specific static information of face images obtained by face analysis technique. Zhu \textit{et al.} \cite{zhu2016deep} utilized the dense correspondence field estimation to help recovering textual details. Meanwhile, some other methods use face image prior knowledge obtained by CNN or GAN-based network. For example, Yu \textit{et al.} \cite{yu2017face, yu2016ultra} adopt a GAN-based network to hallucinate face images. Yu \textit{et al.} \cite{yu2017hallucinating} used a spatial transformer network to address the problem of face scale and misalignment. Chen and Bulat\cite{chen2018fsrnet, bulat2018super} utilized facial geometric priors, such as parsing maps or face landmark heatmaps, to super-resolve LR face images. Moreover, some wavelet-based methods have also been proposed. Huang \textit{et al.} \cite{huang2017wavelet, huang2019wavelet} introduced a method combined with wavelet transformation to predict the corresponding wavelet coefficients. 

\begin{figure*}[ht]
    \centering
    \includegraphics[scale=0.75]{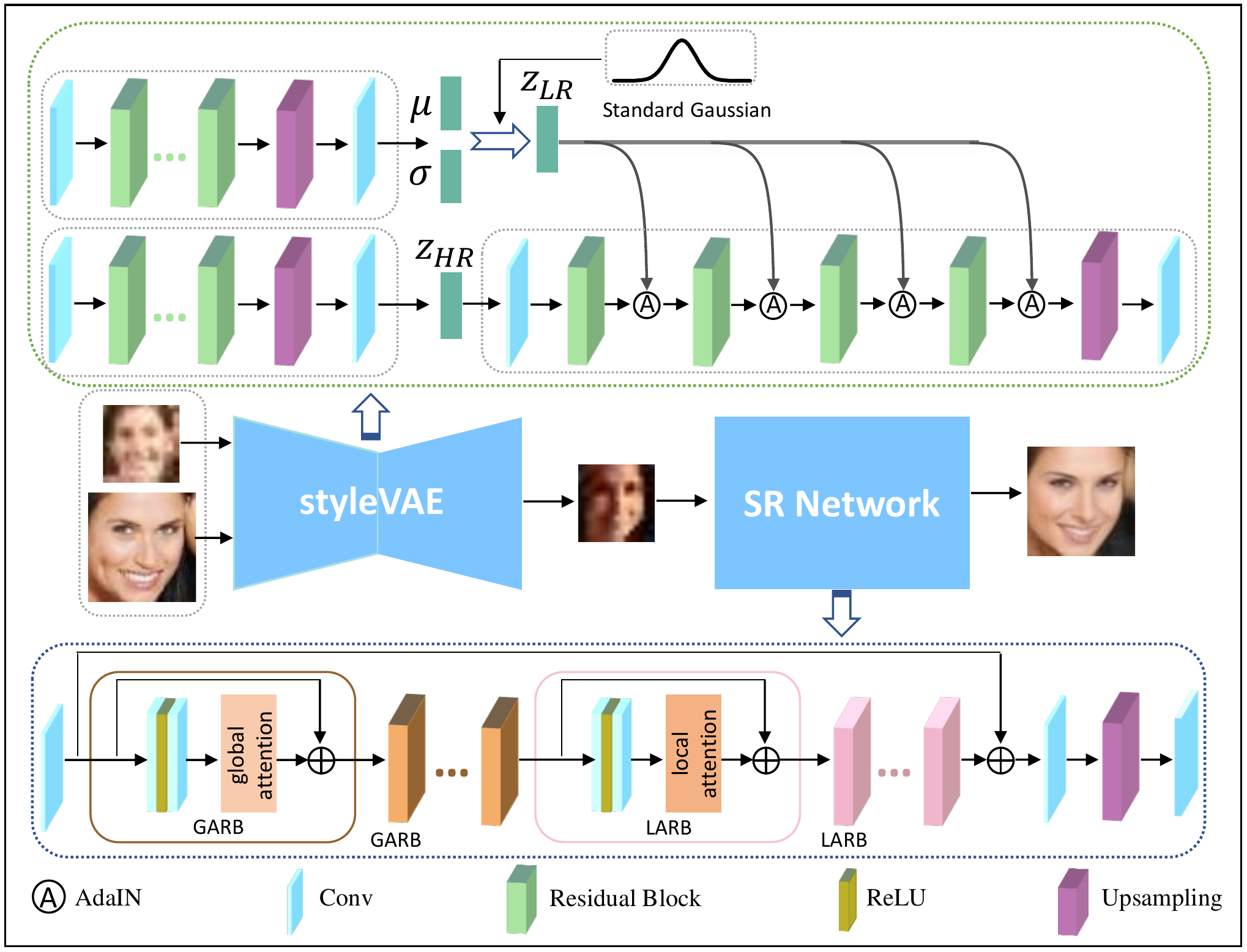}
    \caption{The overall architecture of our proposed method. The proposed styleVAE takes unpaired LR and HR images as its inputs to generate real-world-like LR images. Our proposed SR network takes paired real-world-like LR and HR images as its inputs. By simulating image degradation processes in reality through styleVAE, we can improve the performance of SR methods for real-world SISR.}
    \label{Fig. 2}
\end{figure*}

\textbf{Variational Autoencoder.}
Variational Autoencoder \cite{kingma2013auto} consists of two networks, which are an inference network ($E$) and a generator network ($D$). The inference network ($E$) $q_\phi(z|x)$ encodes the variable $x$ into a latent code $z$ which is supposed to approximate the prior $p(z)$. $D$ samples the variable $x$ from the given latent code $z$. VAE should maximize the variational lower bound (also called evidence lower bound, ELBO):
\begin{equation}
    \begin{split}
        logp_\theta(x) &\geq \mathbb{E}_{q_\phi(z|x)}[logp_\theta(x|z)]
        \\&-D_{KL}(q_\phi(z|x)||p(z)),
    \end{split}
\end{equation}
where the first item denotes the process of reconstructing $x$ from the posterior $q_\phi(z|x)$. The second item means Kullback-Leibler divergence between the prior and the posterior distribution. 

\section{Methodology}
Fig. \ref{Fig. 2} shows the overall architecture of our method that consists of two stages. In the first stage, styleVAE is proposed to generate real-world-like LR images. In the second stage, the generated LR images paired with the corresponding HR images are fed into the SR network for super-resolution. After training, our proposed pipeline can produce pleasing reconstructed images for real-world SISR. In this section, we firstly introduce how to generate real-world-like LR images by using styleVAE and then detail the SR network.

\begin{figure*}[ht]
    \centering
    \includegraphics[scale=0.76]{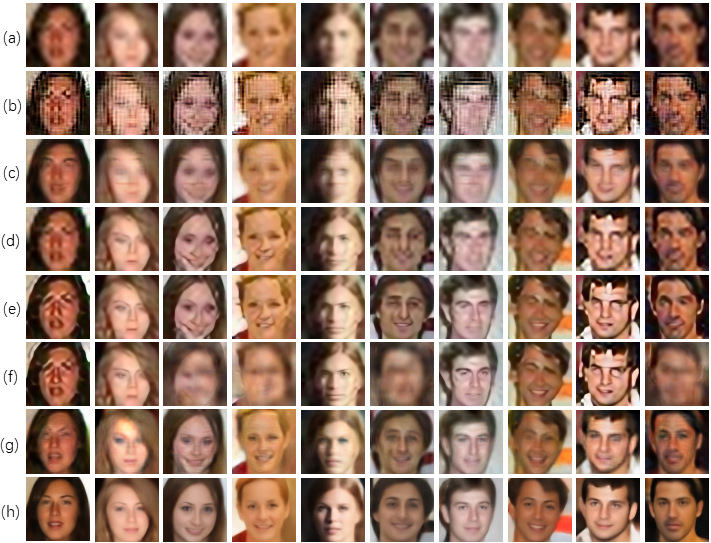}
    \caption{Comparisons with the state-of-art methods ($4\times$). (a) Input real-world LR images. (b) Results of SRCNN \cite{dong2015image}. (c) Results of SRGAN \cite{ledig2017photo}. (d) Results of VDSR \cite{kim2016accurate}. (e) Results of EDSR \cite{Lim_2017_CVPR_Workshops}. (f) Results of RDN \cite{zhang2018residual}. (g) Results of \cite{bulat2018learn}'s method (h) Our results. Compared with other methods, our proposed pipeline can reconstruct sharper SR images with better details.}
    \label{Fig. 3}
\end{figure*}

\subsection{Style Variational Autoencoder}
We now describe how to simulate real-world LR images by using styleVAE in detail. We can get a more stable training process using VAE rather than GAN \cite{goodfellow2014generative}. Moreover, with the help of style transfer, both the style of real-world LR images and the content of HR images are preserved well. To better encode style information, we also maximize the mutual information between the input and the generated images. Some generated examples are shown in Fig \ref{Fig. 7}.

\textbf{Style Transfer.}
There are various nuisance factors in real-world LR images, including motion blur, lens aberration, and sensor noise and so on. We expect to transfer nuisance factors in real-world LR images to generated LR images. We choose Adaptive Instance Normal (AdaIN) to achieve our purpose due to its suitableness, efficiency and compact representation. Fig. \ref{Fig. 2} shows the architecture of our proposed styleVAE, there are two inference networks $E_{LR}$ (the upper left corner) and $E_{HR}$ (the lower left corner), and one generator $G$ (the bottom right corner) in styleVAE. $E_{LR}$ and $E_{HR}$ project input real-world LR images and HR images into two latent spaces, representing style information and content information, respectively. The two latent codes produced by the inference networks $E_{LR}$ and $E_{HR}$ are combined in a style transfer way (AdaIN) rather than concatenated directly. The style information $y$ (i.e., $Z_{LR}$) can control AdaIN \cite{gatys2016image,huang2017arbitrary} operations after each residual block in the generator $G$. The AdaIN can be calculated as follows:
\begin{equation}
    \begin{split}
        AdaIN(x,y)=y_{s}\frac{x-\mu(x)}{\sigma(x)}+y_{b},
    \end{split}
\end{equation}
where $x$ denotes the feature maps from the previous layer. $y_{s}$ and $y_{b}$ are obtained from $y$ through a fully connected layer and used as the scale factor and the bias, respectively. Note that the AdaIN calculation for each feature map $x$ is independent.

Following VAE, we use the Kullback-Leibler (KL) divergence to regularize the latent space obtained by $E_{LR}$. The distribution of the latent space is supposed to approximate Gaussian distribution. Thus, sampling from Gaussian distribution provides us the possibility to produce LR images with diversity. The $E_{LR}$ branch has two output variables, i.e., $\mu$ and $\sigma$. To a reparameterization trick, we have $Z_{LR}={\mu}_{LR}+\sigma\odot\epsilon$, where $\epsilon \sim \mathcal{N}(0,I)$, $\odot$ means Hadamard product; $\mu$ and $\sigma$ denote the mean and the standard deviation, respectively. Given $N$ data samples, the posterior distribution $q_\phi(z|x_{LR})$ is constrained through Kullback-Leibler divergence:
\begin{equation}
    \begin{split}
        \mathcal{L}_{kl} &= KL(q_\phi(z|x_{LR})||p(z))
        \\&= \frac{1}{2N}\sum_{i=1}^N\sum_{j=1}^M(1+log(\sigma_{ij}^2)-\mu_{ji}^2-\sigma_{ij}^2),
    \end{split}
\end{equation}
where $q_\phi(\cdot)$ is the inference network $E_{LR}$. The prior $p(z)$ is the standard multivariate Gaussian distribution. $M$ is the dimension of $z$.

The generator $p_\theta(y_{LR}|z_{LR},z_{HR})$ in styleVAE is required to generate LR images $y_{LR}$ from the latent space $z_{HR}$ and the learned distribution $z_{LR}$. Similar \cite{huang2017arbitrary, huang2019wavelet}, we use a pre-trained VGG network \cite{simonyan2014very} to calculate the following loss function:
\begin{equation}
    \begin{split}
        \mathcal{L}_{style} = \alpha\mathcal{L}_{c} + \beta\mathcal{L}_{s},
    \end{split}
\end{equation}
where $\alpha$ and $\beta$ are the weights for the content loss and the style loss, respectively. Here we set $\alpha$ and $\beta$ to 1 and 0.1, respectively. The content loss function is expected to ensure that generated LR images can retain the content of the corresponding HR images ($x_{HR}$). It defines at a specific layer $J$ of the VGG network \cite{simonyan2014very}:
\begin{equation}
    \begin{split}
        \mathcal{L}_{c} = \left \|\phi^J(y_{LR}) - \phi^J(x_{HR}) \right\|_F^2,
    \end{split}
\end{equation}
where $y_{LR}$ and $x_{HR}$ denote generated the LR images and the corresponding HR images, respectively. We resize the size of $x_{HR}$ to match that of $y_{LR}$. Furthermore, $\mathcal{L}_{s}$ is defined by a weighted sum of the style loss at different layers of the pre-trained VGG network:
\begin{equation}
    \begin{split}
        \mathcal{L}_{s} = \sum_i w_{i}\mathcal{L}^i_{s}(y_{LR}, x_{LR}),
    \end{split}
\end{equation}
where $w_i$ is the trade-off factor for the style loss $\mathcal{L}^i_{s}$ at $i-th$ layer of the pre-trained VGG network. $\mathcal{L}^i_{s}$ is computed as the Euclidean distance between the Gram matrices of the feature maps for $y_{LR}$ and $x_{LR}$. In \cite{li2017demystifying}, this style loss has been studied in detail.

\textbf{Mutual information maximization.} The purpose of the inference network $E_{LR}$ is to extract the style information, which is essential for the subsequent generator. To gain style representation better, the mutual information between real-world and generated LR images is required to be maximized. Inspired by \cite{belghazi2018mine}, we estimate the mutual information between two random high dimensional variables by using the gradient descent over neural networks. It could be mathematically formulated as follows:
\begin{equation}
    \begin{split}
        \mathcal{L}_{mi}&=\sup \limits_{\theta\in\Theta}\mathbb{E}_{p_{(x_{LR}, y_{LR})}}[T_{\theta}]
        \\&-log(\mathbb{E}_{p(x_{LR})\otimes p(x_{HR})}[e^{T_\theta}]),
    \end{split}
\end{equation}
where $T_\theta$ denotes a static deep neural network parameterized by $\theta\in\Theta$. The inputs of the $T_\theta$ are empirically drawn from the joint distribution $p_{(x_{LR}, y_{LR})}$ and the product of the marginal $p_{x_{LR}} \otimes p_{y_{LR}}$. More details can be found in our supplementary materials.

According to all the loss functions mentioned above, the overall loss to optimize styleVAE can be formulated as:
\begin{equation}
    \begin{split}
        \mathcal{L}_{styleVAE} = \lambda_1\mathcal{L}_{kl} + \lambda_2\mathcal{L}_{style} + \lambda_3\mathcal{L}_{mi},
    \end{split}
\end{equation}
where $\lambda_1$, $\lambda_2$, and $\lambda_3$ are the trade-off factors.

\subsection{Super-Resolution Network}
As illustrated in \cite{huang2017beyond}, TP-GAN simultaneously perceives global topology information and local texture information and achieves pleasing results. Different from it, our proposed SR network firstly build global topology structures or sketches by using global attention residual block (GARB) \cite{zhang2018self}. Then our SR network focus on filling local details into coarse SR images through local attention residual block (LARB) \cite{hu2019local}. The SR network accepts LR images produced by styleVAE as inputs and mainly contains 8 GARBs and LARBs, respectively. We train our SR network with $L1$ for better performance instead of $L2$ \cite{zhao2015loss}.

\begin{figure}[ht]
    \centering
    \includegraphics[scale=0.49]{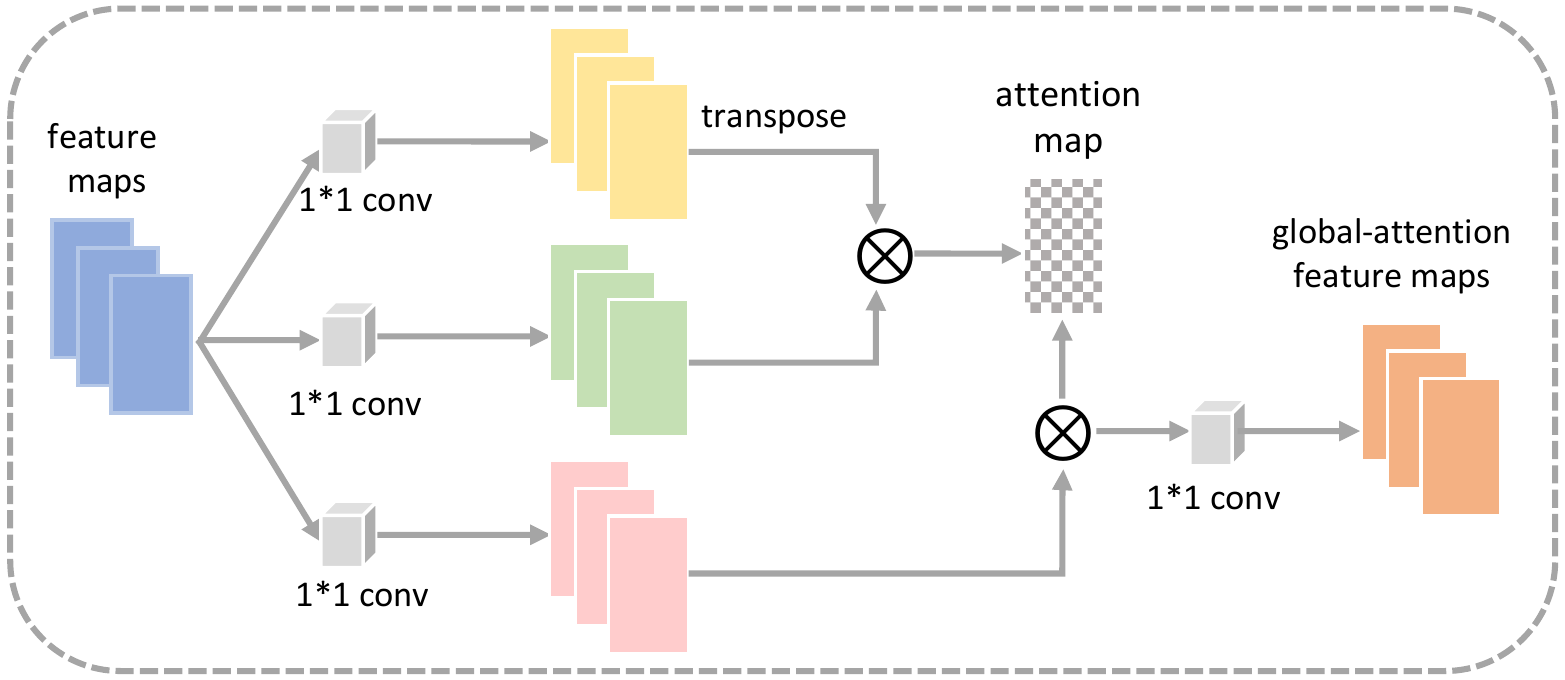}
    \caption{The proposed GA module in GARB. The $\otimes$ denotes the matrix multiplication operation.}
    \label{Fig. 5}
\end{figure}

\begin{figure}[ht]
    \centering
    \includegraphics[scale=0.45]{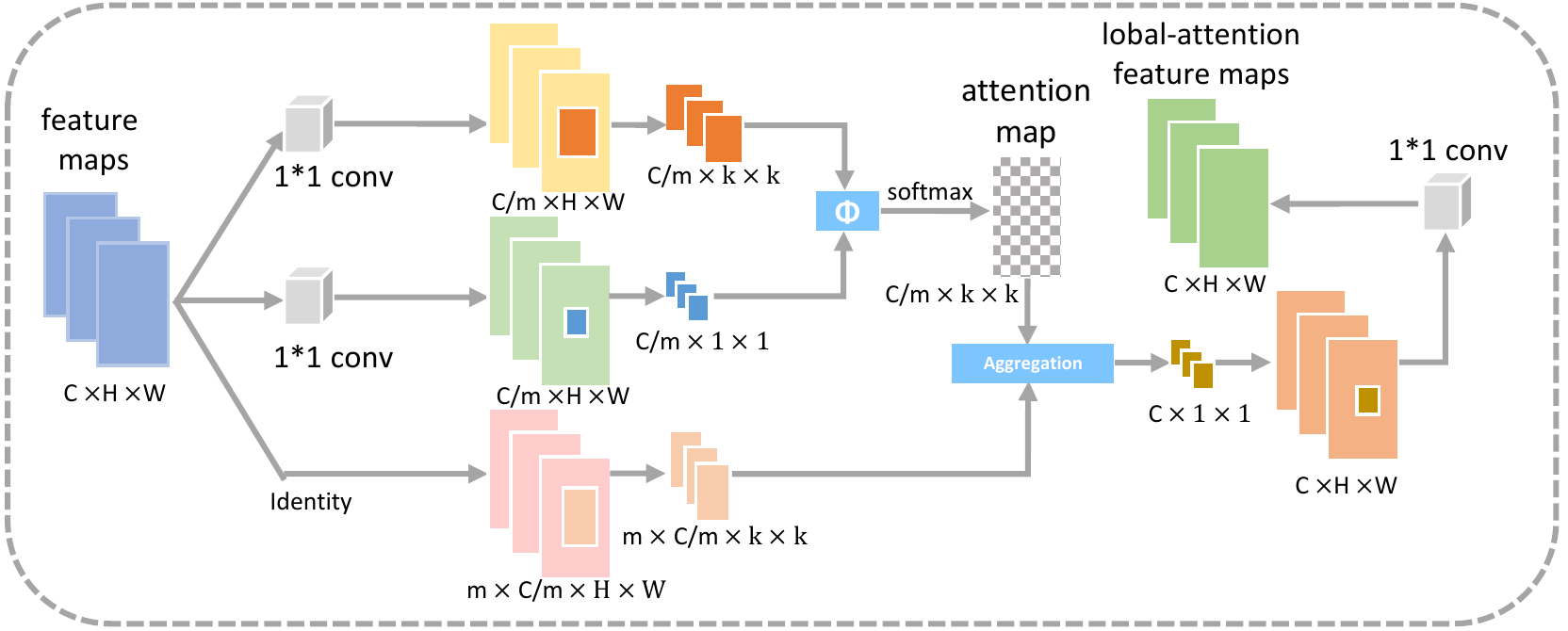}
    \caption{The proposed LA module in LARB. Different from GA, it can obtain the composability between target pixel and a pixel from visible scope of the target pixel instead of all pixel of images.}
    \label{Fig. 6}
\end{figure}

\textbf{Global Attention Residual Block.} We now describe GARB in detail. Most existing models depend heavily on convolution layers to learn the dependencies of different regions of input images for capturing global topology information. But this way leads to incorrect geometric or structural patterns. we propose a global attention residual block to learn long-range dependencies by using global attention (GA) module \cite{zhang2018self,wang2018non}. It can maintain efficiency in calculation and statistics. Fig. \ref{Fig. 5} shows the structure of the GA module. There is a skip connection in GARB due to the success of residual blocks (RBs) \cite{zhang2018residual} (See Fig. \ref{Fig. 2}). The GA module can be formulated as follows:
\begin{equation}
    \begin{split}
        \beta_{j,i}=\frac{exp(s_{i,j})}{\sum_{i=1}^{N}exp(s_{i,j})},
    \end{split}
\end{equation}
where $s_{i,j}=f(x_i)^Tg(x_j)$ ($f(x)=W_fx$, $g(x)=W_gx$) represents that the feature maps of the former hidden layer are projected into two latent spaces to obtain the attention value. $\beta_{i,j}$ indicates the degree of attention that the $i^{th}$ position receives when generating the $j^{th}$ area. The output of the attention layer is defined as:
\begin{equation}
    \begin{split}
        o_j=v(\sum_{i=1}^{N}\beta_{(j,i)}h(x_i)),
    \end{split}
\end{equation}
where $h(x_i)=W_hx_i$, $v(x_i)=W_vx_i$. The above $W_f$, $W_g$, $W_h$, $W_v$ are implemented by a convolution layer with kernel size $1\times1$. We connect $o_i$ and $x_i$ in a residual way, so the final output is shown as below:
\begin{equation}
    y_i = \lambda{o_i} + x_i,
\end{equation}
where $\lambda$ is a learnable scalar.

\begin{figure}[htb]
    \centering
    \includegraphics[scale=0.7]{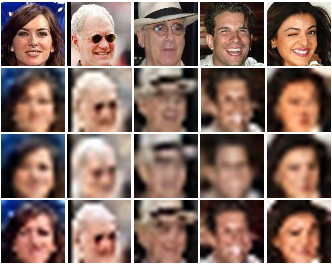}
    \caption{Examples of different LR images generated by styleVAE. The first row is input HR images. The other rows are generated LR images. We discover that styleVAE can produce a variety of real-world-like LR images.}
    \label{Fig. 7}
\end{figure}

\textbf{Local Attention Residual Block.} SISR can be seen as the process of filling local details into coarse SR images. we expect SR networks can generate as more local details as possible. Convolution variants also process aggregation computation in a local neighborhood. But the performance of them are usually saturated with the size of neighborhoods larger than $3\times3$. Thus we propose a local attention residual block (LARB) to capture local details though local attention (LA) module \cite{hu2019local}. The LA module forms local pixel pairs with a flexible bottom-up way, which can efficiently deal with visual patterns with increasing size and complexity. The structure of LARB and the LA module are shown in Fig. \ref{Fig. 2} and Fig. \ref{Fig. 6}, respectively. We use a general method of relational modeling to calculate the LA module and can be defined as:
\begin{equation}
    \begin{split}
        \omega(p^{'},p)=softmax(\Phi(f_{\theta_q}(x_{p^{'}}),f_{\theta_k}(x_p))),
    \end{split}
\end{equation}
where $\omega(p^{'},p)$ obtains a representation at one pixel by computing the composability between it (target pixel $p^{'}$) and a pixel $p$ in its visible position range. Transformation functions $f_{\theta_q}$ and $f_{\theta_k}$ are implemented by $1\times1$ convolution layer. The function $\Phi$ is chosen the squared difference: 
\begin{equation}
    \begin{split}
        \Phi(q_{p^{'}},k_{p})= -(q_{p^{'}}-k_{p})^{2}.
    \end{split}
\end{equation}

\section{Experimental Results}
In this section, we firstly introduce the datasets and implementation in detail. Then we evaluate our proposed method from both qualitative and quantitative aspects.

\subsection{Datasets and Implementation}
\textbf{Training dataset.} As illustrated in \cite{bulat2018learn}, we select the following four datasets to build the HR training dataset that contains 180k faces. The first is a subset of VGGFace2 \cite{Cao18} that contains images with 10 large poses for each identity (9k identities). The second is a subset of Celeb-A \cite{liu2015deep} that contains 60k faces. The third is the whole AFLW \cite{koestinger2011annotated} that contains 25k faces used for facial landmark localization originally. The last is a subset of LS3D-W \cite{bulat2017far} that contains face images with various poses.

We also utilize the WIDER FACE \cite{yang2016wider} to build a real-world LR dataset. WIDER FACE is a face detection benchmark dataset that consists of more than 32k images affected by various of noise and degradation types. We randomly select 90\% images in the LR training dataset.

\textbf{Testing dataset.} Another 10\% images from WIDER FACE described in the latest section are selected as real-world LR testing dataset. We conduct experiments on it to verify the performance of our proposed method. 

\textbf{Implementation Details.} Our proposed styleVAE is trained on the unpaired training HR and LR images for 10 epochs. After that, the paired LR-HR dataset is created used to train the SR network with 50 epochs. We train our styleVAE and SR network through the ADAM algorithm \cite{kingma2014adam} with $\beta_1=0.9$, $\beta_2=0.999$. The learning rate is initially set to $10^{-4}$ and remains unchanged during the training process. The weights $\lambda_{1}$, $\lambda_{2}$ and $\lambda_{3}$ are set as 0.01, 1 and 0.1, respectively. We use PyTorch to implement our models and train them on NVIDIA Titan Xp GPUs.

\begin{table}[!t]
\begin{center}
\begin{tabular}{|l|c|c|c|}
\hline
METHOD & PSNR  & SSIM   & FID    \\
\hline\hline
BICUBIC & 21.19 & 0.5570  & 288.47 \\
SRCNN \cite{dong2015image}  & 19.57 & 0.4030 & 256.78 \\
SRGAN \cite{ledig2017photo} & 20.36 & 0.5007 & 179.70 \\
VDSR \cite{kim2016accurate}  & 20.61 & 0.5584 & 144.29 \\
EDSR \cite{Lim_2017_CVPR_Workshops}  & 20.44 & 0.5137 & 129.93 \\
RDN \cite{zhang2018residual}   & 18.18 & 0.4063 & 162.04 \\
Bulat's \cite{bulat2018learn} & 22.76 & 0.6296 & 149.97 \\
OURS   & \textbf{24.16} & \textbf{0.7197} & \textbf{98.61} \\
\hline
\end{tabular}
\end{center}
\caption{The second and third columns show PSNR and SSIM based performance on synthetic real-world-like LR images (Higher is better). The fourth column shows FID based performance on our testing dataset (Lower is better).}
\label{Tab. 1}
\end{table}

\subsection{Experiments on Real-World Images}
In this section, We conduct experiments on the testing dataset described in Section 4.1. In order to evaluate the performance of our proposed method, we compare with the following state-of-the-art methods both numerically and qualitatively: SRCNN \cite{dong2015image}, SRGAN \cite{ledig2017photo}, VDSR \cite{kim2016accurate}, EDSR \cite{Lim_2017_CVPR_Workshops}, Bulat's method \cite{bulat2018learn} and RDN \cite{zhang2018image}. We retrain all these compared methods for the sake of fairness on our HR training dataset with the default configurations described in their respective papers. Note that LR images are produced by applying a bicubic kernel to the corresponding HR images.


In numerical terms, we use Fr\'{e}chet Inception Distance (FID) \cite{heusel2017gans} to measure the quality of the generated images since there are no corresponding HR images. The quantitative results of different SR methods on our testing dataset are summarized in Table \ref{Tab. 1} (with the factor $\times 4$). It clearly demonstrates that our proposed method is superior to other compared approaches and achieves the best performance on the testing dataset. We also discover that the performances of compared methods trained on bicubic-downsampled LR images are degraded when applied to real-world LR images. The main reason is that nuisance factors, e.g. motion blur, lens aberration and sensor noise, are not taken into synthetic LR images by bicubic interpolation. By training on real-world-like LR images, our method is superior to all compared methods, and the FID value is reduced by a maximum of 158.17.

In Fig. \ref{Fig. 3}, we visually show the qualitative results the our testing dataset with $\times 4$ scale. There are significant artifacts in HR images generated by shallower networks, e.g. SRCNN \cite{dong2015image} and SRGAN \cite{ledig2017photo}. Serious mesh phenomenons are found in reconstructed images by SRCNN. We can also discover that generated images of VDSR \cite{kim2016accurate}, EDSR \cite{Lim_2017_CVPR_Workshops} and RDN \cite{zhang2018residual} are usually distorted. On the contrary, SR images generated by our proposed method are more realistic than Bulat's method \cite{bulat2018learn}, since LR images produced by styleVAE exceedingly resemble real-world LR images.

\subsection{Experiments on Real-World-Like Images}
In order to verify the performance of the proposed method on LR images with unknown degradation modes, we conduct experiment on synthetic real-world-like LR images obtained by styleVAE with $\times 4$ scale. We utilize images from LFW \cite{LFWTech,LFWTechUpdate, Huang2007a, Huang2012a} as the HR image inputs of styleVAE to generate real-world-like LR images. The second and third columns of Table \ref{Tab. 1} report PSNR and SSIM results of different SR methods. We find that the performances of compared methods are very limited, even lower than that of bicubic up-sampling directly. It also demonstrates that simulating real-world-like LR images is an effective way to improve performance when applied to real-world LR images. 

\begin{figure}[htb]
    \centering
    \includegraphics[scale=0.7]{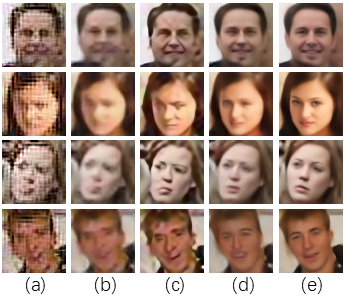}
    \caption{Results of experiments on deep plug-and-play super-resolution. (a) Results of SRCNN-B. (b) Results of SRCNN-S. (c) Results of EDSR-B. (d) Results of EDSR-S. (e) Results of our method. This suggests that the performance of SR methods can be improved by training on LR images generated by styleVAE.}
    \label{Fig. 8}
\end{figure}

\subsection{Experiments on deep Plug-and-Play Super-Resolution}
To further validate the effectiveness of styleVAE, we design two pipelines with the help of plug-and-play framework. We can simply plug styleVAE into SR networks to replace bicubic down-sampled LR images that are used in many previous SR methods. We choose two of the compared methods as the plugged SR networks: a shallower SR network SRCNN \cite{dong2015image} and a deeper SR network EDSR. Thus there are four versions of SR networks: SRCNN-B and EDSR-B, trained on bicubic down-sampled LR images, SRCNN-S and EDSR-S, trained on LR images generated by styleVAE. The FID results on our testing dataset are reported in Table \ref{Tab. 3}. As can be seen from Table \ref{Tab. 3}, the FID values of SRCNN-B and EDSR-B (the second row of Table \ref{Tab. 3}) are higher than those of EDSR-S and EDSR-S (the last row of the Table \ref{Tab. 3}). By simulating real-world LR images using styleVAE, SRCNN can gain an improvement 58.3 (the third column of Table \ref{Tab. 3}) and EDSR can be improved by 22.3 (the last column of Table \ref{Tab. 3}). 

\begin{table}[!t]
\begin{center}
\begin{tabular}{|c|c|c|c|}
\hline
DATA TYPE & SCALE              & SRCNN  & EDSR   \\ 
\hline\hline
BICUBIC   &\multirow{2}{*}{$\times4$} & 256.78 & 129.93 \\
styleVAE  &                    & 198.75 & 107.63       \\ 
\hline
\end{tabular}
\end{center}
\caption{Results of experiments on deep plug-and-play SR in FID. (Lower is better).}
\label{Tab. 3}
\end{table}

We also demonstrate the visual results in Fig. \ref{Fig. 8}. As shown in Fig. \ref{Fig. 8}, compared (a) with (b), SRCNN-S can effectively eliminate the mesh phenomenon in the image generated by SRCNN-B. When training on LR images generated by styleVAE, EDSR-S can produce more pleasing results (d) rather than distorted reconstructed images (c) by EDSR-B. Compared (b), (d) and (e), our proposed method is able to generate sharper images. 

\subsection{Ablation studies}
\textbf{StyleVAE.} In order to investigate the effectiveness of the mutual information estimation (MI) and Adaptive Instance Normalization (AdaIN) used in styleVAE, we train several other variants of styleVAE: remove MI or/and AdaIN. To evaluate the performance of these variants of styleVAE, we measure FID between LR images generated by these variants and real-world LR images from our testing dataset. FID results are provided in Table \ref{tab.6}. When both AdaIN and MI are removed, the FID value is relatively high. After arbitrarily adding one of the two, the value of FID is decreased. For both MI and AdaIN used in styleVAE, the FID result is the lowest. We also evaluate how similar the synthetic LR images by bicubic down-sample and real-world LR images from WIDER FACE. The FID result is found as 31.20. These results faithfully indicate that AdaIN and MI are essential for styleVAE. 

\textbf{SR network.} Similar to the ablation investigation of styleVAE, we also train several variants of the proposed SR network: remove the GA or/and LA module(s) in the SR network. These several variants are trained on LR images produced by performing bicubic interpolation on corresponding HR images. In Table \ref{Tab. 4}, when both the GA and LA modules are removed, the PSNR value on LFW (with upscale factor $\times 4$) is the lowest. When the LA module is added, the PSNR value is increased by 0.1 dB. After adding the GA module, the performance reaches 30.27 dB. When both attention modules are added, the performance is the best, with a PSNR of 30.43 dB. These experimental results clearly demonstrate that these two attention modules are necessary for the proposed SR network and can greatly improve its performance.

\begin{table}[!t]
\begin{center}
\begin{tabular}{ |c|c|c|c|c| }
\hline
GA&  $\times$&  $\times$& $\checkmark$& $\checkmark$ \\
LA&  $\times$&  $\checkmark$&  $\times$&  $\checkmark$ \\
\hline\hline
PSNR& 30.08& 30.18&  30.27&  30.43 \\
\hline
\end{tabular}
\end{center}
\caption{Ablation investigation on the effects of the GA and LA modules in SR network. The PSNR (dB) values are reported on LFW (higher is better).}
\label{Tab. 4}
\end{table}

\begin{table}[!t]
\begin{center}
\begin{tabular}{|c|c|c|c|c|}
\hline
AdaIN&  $\times$&  $\times$& $\checkmark$& $\checkmark$ \\
MI   &  $\times$&  $\checkmark$&  $\times$&  $\checkmark$ \\
\hline\hline
FID& 26.6& 25.78& 24.63 &18.77\\
\hline
\end{tabular}
\end{center}
\caption{Investigations of AdaIN and MI in styleVAE. We also evaluate how similar the synthetic LR images by bicubic down-sample and real-world LR images in WIDER FACE. The FID value between these is 31.20 (Lower is better).}
\label{tab.6}
\end{table}

\section{Conclusion}
We propose a novel two-stage process to address the challenging problem of super-resolving real-world LR images. The proposed pipeline unifies a style-based Variational Autoencoder (styleVAE) and an SR network. Due to the participation of nuisance factors transfer and VAE, styleVAE can generate real-world-like LR images. Then the generated LR images paired with the corresponding HR images are fed into the SR network. Our SR network firstly learns long-range dependencies by using GARB. Then the attention of SR network moves to local areas of images in which texture detail will be filled out through LARB. Extensive experiments show our superiority over existing state-of-the-art SR methods and the ability of styleVAE to facilitate method generalization and robustness to real-world cases.

\bibliographystyle{IEEEtran}
\bibliography{egbib}

\end{document}


\title{Supplementary materials for: Exploiting Style and Attention in Real-World Super-Resolution}

\author{\IEEEauthorblockN{Xin Ma$^{1,2}$, Yi Li$^{1,2}$, Huaibo Huang$^{1,2}$, Mandi Luo$^{1,2}$, Ran He$^{1,2}$}
\IEEEauthorblockA{$^{1}$NLPR$\;\&\;$CEBSIT, CASIA \quad
{$^2$}University of Chinese Academy of Sciences\\
{\tt\small \{xin.ma, yi.li, huaibo.huang\}@cripac.ia.ac.cn, luomandi2019@ia.ac.cn, rhe@nlpr.ia.ac.cn}}
}

\maketitle

\section{Network Architecture}
Table 1 and 2 show the architectures of $E_{LR}$ and $E_{HR}$ in StyleVAE. Table 3 shows the architecture of $G$ in StyleVAE. Table 4 shows the architecture of SR network.
\begin{table}[htb]
\begin{center}
\begin{tabular}{c||c}
\hline
Conv\_stem    & conv{[}64,3,1,1{]}                                                                                             \\ 
\hline
Conv\_body    & \begin{tabular}[c]{@{}l@{}}\{conv{[}64,3,1,1{]},\\ BN,\\ LeakyReLU,\\ conv{[}64,3,1,1{]}\\ BN\}*8\end{tabular} \\ 
\hline
Downsample & \begin{tabular}[c]{@{}l@{}}\{deconv{[}64,4,2,1{]},\\ BN,\\ LeakyReLU\}*2\end{tabular}                          \\ 
\hline
Conv\_last    & conv{[}256,3,1,1{]}                                                                                            \\ 
\hline
\end{tabular}  
\end{center}
\caption{The architecture of $E_{LR}$ in StyleVAE. Conv\_stem, Conv\_body, Downsample and Conv\_last denote layer names. "conv${[}64,3,1,1{]}$" denotes a convolution layer with filters, kernel, stride, padding, which are 64, 3, 1, 1, respectively. "deconv[64,4,2,1]" denotes a deconvolution layer with filters, kernel, stride, padding, which are 64,4,2,1. $\{*\}*n$ denotes n cascaded base blocks.}
\end{table}

\begin{table}[htb]
\begin{center}
\begin{tabular}{c||c}
\hline
Conv\_stem    & conv{[}64,3,1,1{]}                                                                                             \\ 
\hline
Conv\_body    & \begin{tabular}[c]{@{}c@{}}\{conv{[}64,3,1,1{]},\\ BN,\\ LeakyReLU,\\ conv{[}64,3,1,1{]}\\ BN\}*8\end{tabular} \\ 
\hline
Downsample & \begin{tabular}[c]{@{}c@{}}\{deconv{[}64,4,2,1{]},\\ BN,\\ LeakyReLU\}*4\end{tabular}                          \\ 
\hline
Conv\_last    & conv{[}256,3,1,1{]}                                                                                            \\ 
\hline
\end{tabular}
\end{center}
\caption{The architecture of $E_{HR}$ in styleVAE.}
\end{table}

\begin{table}[htb]
\begin{center}
\begin{tabular}{c||c}
\hline
Conv\_stem   & conv{[}256,3,1,1{]}                                                                                                          \\ \hline
Res\_1       & \begin{tabular}[c]{@{}c@{}}\{conv{[}64,3,1,1{]},\\ BN,\\ LeakyReLU,\\ conv{[}64,3,1,1{]},\\ BN,\\ \}*2,\\ AdaIN\end{tabular} \\ \hline
Res\_2       & \begin{tabular}[c]{@{}c@{}}\{conv{[}64,3,1,1{]},\\ BN,\\ LeakyReLU,\\ conv{[}64,3,1,1{]},\\ BN,\\ \}*2,\\ AdaIN\end{tabular} \\ \hline
Res\_3       & \begin{tabular}[c]{@{}c@{}}\{conv{[}64,3,1,1{]},\\ BN,\\ LeakyReLU,\\ conv{[}64,3,1,1{]},\\ BN,\\ \}*2,\\ AdaIN\end{tabular} \\ \hline
Res\_4       & \begin{tabular}[c]{@{}c@{}}\{conv{[}64,3,1,1{]},\\ BN,\\ LeakyReLU,\\ conv{[}64,3,1,1{]},\\ BN,\\ \}*2,\\ AdaIN\end{tabular} \\ \hline
Conv\_middle & conv{[}64,3,1,1{]}                                                                                                           \\ \hline
Upsample     & PixelShuffle                                                                                                                 \\ \hline
Conv\_last   & conv{[}3,3,1,1{]}                                                                                                            \\ \hline
\end{tabular}
\end{center}
\caption{The architecture of generator $G$ in styleVAE. Res\_n, Conv\_middle and Upsample denote layer names.}
\end{table}

\begin{table}[htb]
\begin{center}
\begin{tabular}{c||c}
\hline
Conv\_stem   & conv{[}64,3,1,1{]}                                                                                                      \\ \hline
GARN         & \begin{tabular}[c]{@{}c@{}}\{conv{[}64,3,1,1{]},\\ LeakyReLU,\\ conv{[}64,3,1,1{]},\\ global attention\}*8\end{tabular} \\ \hline
LARN          & \begin{tabular}[c]{@{}c@{}}\{conv{[}64,3,1,1{]},\\ LeakyReLU,\\ conv{[}64,3,1,1{]},\\ local attention\}*8\end{tabular}  \\ \hline
Conv\_Middle & conv{[}64,3,1,1{]}                                                                                                      \\ \hline
Upsample     & PixelShuffle                                                                                                            \\ \hline
Conv\_last   & conv{[}3,3,1,1{]}                                                                                                       \\ \hline
\end{tabular}    
\end{center}
\caption{The architecture of SR network. GARN and LARN denote global attention residual network and local attention residual network, respectively.}
\end{table}

\section{Algorithm for Mutual information maximization}
As described in section Methodology, mutual information is used to extract the style information. We maximize the mutual information between the input low-resolution face images and the generated low-resolution images. Followed by \cite{belghazi2018mine}, given $\hat{P}^{(n)}$ as the empirical distribution associated with $n$ samples, the objective function can be defined as,
\begin{equation}
    \begin{split}
        \widehat{I(x_{LR}, y_{LR})}_n&=\sup \limits_{\theta\in\Theta}\mathbb{E}_{P_{(x_{LR}, y_{LR})}^{(n)}}[T_{\theta}]
        \\&-log(\mathbb{E}_{p(x_{LR}^{(n)})\otimes \hat{p}_{(x_{HR})}^{(n)}}[e^{T_\theta}]),
    \end{split}
\end{equation}
where $T_\theta$ denotes a static deep neural network parameterized by $\theta\in\Theta$. The inputs of the $T_\theta$ are empirically sampled from the joint distribution $p_{(x_{LR}, y_{LR})}$ and the product of marginals $p_{x_{LR}} \otimes p_{y_{LR}}$.

The optimization algorithm of the estimation network $T_{\theta}$ is provided in Algorithm 1.

\begin{algorithm}[h]
\caption{Maximization mutual information}
\begin{algorithmic}[1]
\STATE $\theta$ $\leftarrow$ initialize network parameters;
\REPEAT
\STATE Take $b$ minibatch samples from the joint distribution:
\STATE $(x_{LR}^{(1)},y_{LR}^{(1)}),...,(x_{LR}^{(b)},y_{LR}^{(b)})\sim P_{(x_{LR},y_{LR})}$
\STATE Take $n$ samples from the $y_{LR}$ marginal distribution:
\STATE $\overline{y}^{(1)}_{LR},...,\overline{y}^{(b)}_{LR}\sim P_{(y_{LR})}$
\STATE Evaluate the lower-bound:
\STATE $\nu(\theta) \leftarrow \frac{1}{b}\sum^{b}_{i=1}T_{\theta}(x_{LR}^{(i)},y_{LR}^{(i)})-log(\frac{1}{b}\sum^b_{i=1}e^{T_{\theta}(x_{LR}^{(i)},\overline{y}_{LR}^{(i)})})$
\STATE Evaluate bias modified gradients:
\STATE $\hat{G}_{\theta} \leftarrow \hat{\nabla}_{\theta}\nu(\theta)$
\STATE Update the parameters of the statistics network:
\STATE $\theta \leftarrow \theta + \hat{G}(\theta)$
\UNTIL{convergence}
\end{algorithmic}
\end{algorithm}

\subsection{The architecture of the static network architecture}
\begin{table}[htb]
\begin{center}
\begin{tabular}{c||c}
\hline
Conv\_1 & conv[16,3,1,0] \\ \hline
Conv\_2 & conv[32,3,1,0] \\ \hline
Conv\_3 & conv[64,3,1,0] \\ \hline
FC\_1   & 1024           \\ \hline
FC\_2   & 1              \\ \hline
\end{tabular}
\end{center}
\caption{The architecture of the static network. FC\_n denotes fully connected layer.}
\end{table}

\bibliographystyle{IEEEtran}
\bibliography{egbib}
